# Smart Metering System Capable of Anomaly Detection by Bi-directional LSTM Autoencoder

Sangkeum Lee
Environment ICT Research Section
Electronics and Telecommunications Research Institute (ETRI)
Daejeon 34141, South Korea
sangkeum@etri.re.kr

Hojun Jin
The Cho Chun Shik Graduate School of Green Transportation
Korea Advanced Institute of Science and Technology (KAIST)
Daejeon 34141, South Korea
hjjin1995@kaist.ac.kr

Sarvar Hussain Nengroo
The Cho Chun Shik Graduate School of Green Transportation
Korea Advanced Institute of Science and Technology (KAIST)
Daejeon 34141, South Korea
sarvar@kaist.ac.kr

Yoonmee Doh
Environment ICT Research Section
Electronics and Telecommunications Research Institute (ETRI)
Daejeon 34141, South Korea
ydoh@etri.re.kr

Chungho Lee
Environment ICT Research Section
Electronics and Telecommunications Research Institute (ETRI)
Daejeon 34141, South Korea
leech@etri.re.kr

Taewook Heo
Environment ICT Research Section
Electronics and Telecommunications Research Institute (ETRI)
Daejeon 34141, South Korea
htw398@etri.re.kr

Dongsoo Har
The Cho Chun Shik Graduate School of Green Transportation
Korea Advanced Institute of Science and Technology (KAIST)
Daejeon 34141, South Korea
dshar@kaist.ac.kr

*Abstract*— Anomaly detection is concerned with a wide range of applications such as fault detection, system monitoring, and event detection. Identifying anomalies from metering data obtained from smart metering system is a critical task to enhance reliability, stability, and efficiency of the power system. This paper presents an anomaly detection process to find outliers observed in the smart metering system. In the proposed approach, bi-directional long short-term memory (BiLSTM) based autoencoder is used and finds the anomalous data point. It calculates the reconstruction error through autoencoder with the non-anomalous data, and the outliers to be classified as anomalies are separated from the non-anomalous data by predefined threshold. Anomaly detection method based on the BiLSTM autoencoder is tested with the metering data corresponding to 4 types of energy sources electricity/water/heating/hot water collected from 985 households.

## I. Introduction

Electric power consumption is increasing rapidly in urban areas due to the change of industry paradigm to the 4th industrial revolution, which requires increased power consumption. Therefore, more power facilities are newly established and the capacities of existing power facilities are being expanded. For instance, about 80% of the existing power facilities are expected to be re-established to satisfy the increasing load demand in Ontario [1]. Generating more electricity in proportion to the increasing power consumption is helpful to alleviate this situation. However, demand management can be more practical solution for a long-term perspective to fully utilize the existing resources and reduce the gap between energy supply and demand. With efficient demand management, efficient power usage can be achieved, and thus the time until the maximum allowed demand occurs can be prolonged.

As modern society relies heavily on the utilization of electricity, power outages have had a great impact on our lives. It is essential to manage the power system efficiently to provide reliable service to the customers. To solve this problem, smart meters are utilized to provide the users with power usage pattern, high frequency of power usage information and bi-directional communication between smart meters and meter data management system (MDMS), and establishing a more trustworthy power system [2]. The smart meter is typically used as power measurement device which records the amount of electricity usage and transfers this information to the MDMS for data processing and storage [3, 4]. To maintain interoperability between the smart meter and the MDMS, standardization of data collection, process, storage, and communication has been followed [5]. As a result, the smart metering allows isolation, restoration, and fault detection in power system with various algorithms [6, 7]. When needed, the bi-directional communication between utility and smart meters can be realized by adopting wireless sensor networks that might require the capability of spectrum sensing [8] and broadband transmission such as OFDM signal transmission [9]. Operation of wireless rechargeable sensor networks has been addressed in various aspects that include finding optimal paths corresponding to minimum total charging time [10, 11].

Anomaly, also called as outlier, represents a data point which is considerably different from the other data. An outlier is an observation that significantly differs from the others, raising concerns about whether it was produced by a distinct cause [12]. In Fig. 1, anomalies in datasets of electricity/water/heating/hot water are presented.

As shown in Fig. 1, the anomalies can be observed as the deviation of abnormal data from the normal ones. In case of sample data including outliers, there exist several characteristics such as large gaps between inlying and outlying observation data, and the deviation between the outliers and the inlier group measured on a standardized scale. For example, the energy consumption of buildings often exceeds the specific value representing energy savings due to anomaly even though the energy model of the building is designed as a low-energy model [13]. To figure out these circumstances, anomaly detection is a helpful method to identify anomalous events. Moreover, analysis and detection of anomalies in power system are important as they have useful information on data generation process. The existing techniques for anomaly detection usually solve specific problems, utilizing various disciplines, such as deep learning techniques, statistical methods, and information theory.

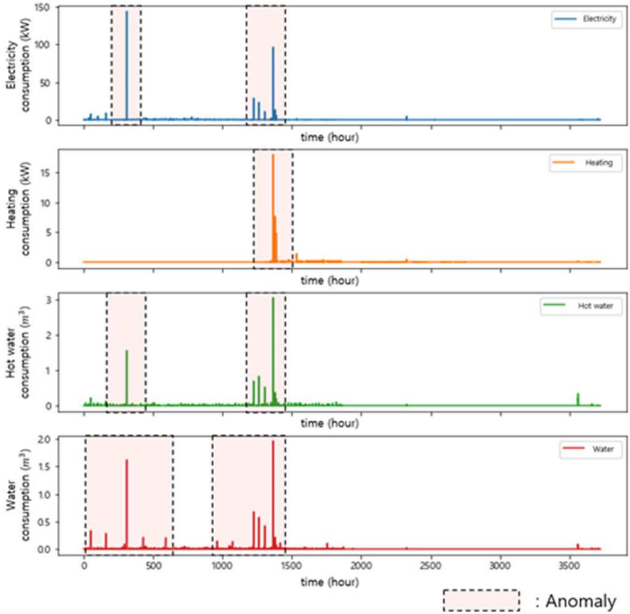

Fig. 1. Anomaly (peak waveform) observed in smart metering dataset.

The setup for the use of deep learning techniques to execute the anomaly detection is fundamentally determined depending on the label applicable to the dataset and therefore it mainly consists of supervised learning, semi-supervised learning, and unsupervised learning. Supervised anomaly detection can be used if the dataset is composed of fully labeled dataset. The more abnormal samples exist in the dataset, the higher performance can be achieved. However, in industrial facilities where anomaly detection is applied, the class-imbalance problem arises as the occurrence rate of abnormal samples is less than that of normal samples. To resolve this issue, different techniques such as data augmentation, loss function design, and batch sampling can be used. In the case of semi-supervised anomaly detection, training data includes only normal data without any anomalies. This methodology is basically to set up a discriminative boundary containing normal samples and narrow it down while judging all samples outside the boundary as abnormal. Unsupervised anomaly detection scores the data focusing only on intrinsic characteristics of the dataset. Compared with supervised anomaly detection, unsupervised anomaly detection can be more useful when analyzing smart metering data in the real world, as labels are not provided for anomalies. Abnormal samples can be found by reducing and restoring dimensions using principal component analysis for given data. Among the neural network based methodologies, an autoencoder based methodology is mainly used, as it is a very powerful tool for high-level data representations. The autoencoder based methodology proceeds with encoding, which compresses the input into the latent variable, and decoding, which restores it close to the original dataset, allowing the autoencoder to learn only significant information in the data. Because the autoencoder does not have dedicated targets, it is trained in an unsupervised method to learn data presentations. In [14], successful anomaly detection of LSTM based autoencoder is presented. An autoencoder could successfully learn to reconstruct time-series data, and the reconstruction error is used to find anomalies. An autoencoder based ensemble method for anomaly detection is described in [15] with different architectures and training schemes. For assessing the performance of anomaly detection algorithms, an attention mechanism in combination with a GRU-based encoder-decoder mechanism is suggested in [16].

In this paper, a deep learning based anomaly detection method is presented to find specific patterns of anomaly in data streams of smart meters. Different studies have dealt with the anomaly detection by using deep learning techniques, considering different situations. However, few works [17, 18] have attempted to get improved performance of anomaly detection. From this perspective, a bi-directional LSTM (BiLSTM) network is applied to autoencoder to get higher accuracy of anomaly detection in the smart metering system. More specifically, the BiLSTM autoencoder based approach is applied to smart metering data composed of 4 types of energy sources electricity/water/heating/hot water collected from 985 apartment households. The BiLSTM network consisting of two uni-directional LSTM networks, takes the input in two ways, one uni-directional LSTM network receiving temporal data from past to future and the other uni-directional LSTM network receiving temporal data from future to past. In this way, the BiLSTM autoencoder can extract the features of temporal data more efficiently, and therefore achieving improved anomaly detection.

The organization of this paper is as follows. In Section II, the structure of the smart metering system is presented. Section III presents the anomaly detection by using the BiLSTM autoencoder. Section IV describes simulation results of anomaly detection for smart metering data. In Section V, concluding remarks are presented.

## II. SMART METERING SYSTEM

A smart meter is a metering device that can measure and transmit real-time energy consumption information to utility providers, facilitating monitoring and control of power system. It is different from the traditional meter in that it can provide a wide range of advanced services, such as remote connection activation/deactivation and bi-directional communication between smart meters and utility providers. Also, the analysis of real-time power usage according to user's behavior can be achieved by smart meters. The overall structure of the smart metering system is made up of different components offering the intended services to the users. In Fig. 2, the overall structure of smart metering system is presented.

Data concentrator is a tool that is commonly found in substations and is used to manage the data collected from various smart meters in distant houses. The data concentrator primarily serves as store-and-forward link between the smart meter and the rest of the system, collecting data on energy expenditure at remote residences, transmitting the data to control centers, and providing the data to the billing system. Furthermore, data concentrators can uncover and organize freshly fitted smart meters, as well as create repeating chains, if necessary. Generally, data concentrators administer a part of infrastructure operations autonomously, such as continuously monitoring the electricity grid and smart meters, confirming failures and disruptions [19], and identifying and investigating theft and interfering attempts. Although data concentrator is unable to take incoming calls, it is capable of initiating and maintaining communication if the connection is lost.

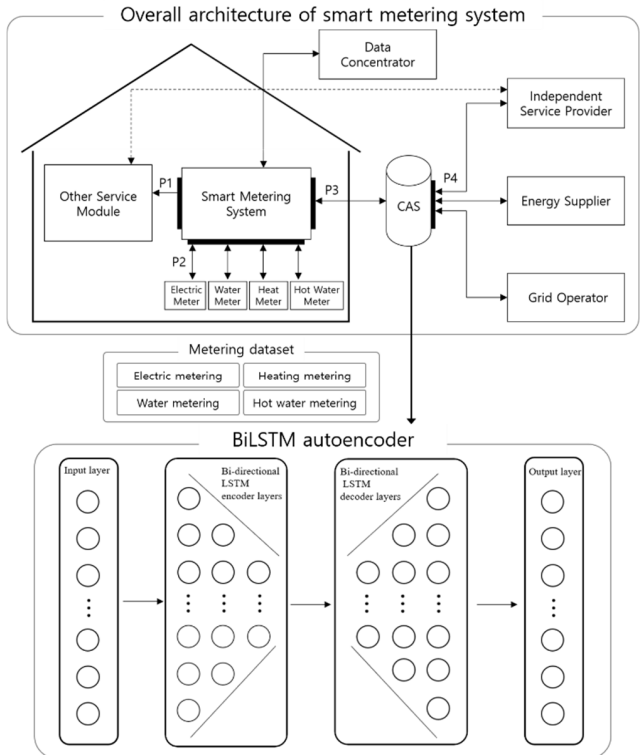

Fig. 2. Overall architecture of smart metering system with BiLSTM autoencoder for anomaly detection.

A central access server (CAS) is described as the central application which manages the information assortment, control, and parameterization, mentioned in NTA 8130. Also, it is used as the centralized authorization for accessing the metering system. Every grid operator keeps a group of servers as a component of the metering framework activity. These servers likewise have diverse programming and applications which are fundamental for the system activities. Also, these servers serve as online portals, allowing clients to access their profiles, for example, check their energy consumption trends. In addition, the metering dataset provided from the servers can be used as the input, and the output of the autoencoder for its training, as presented in Fig. 2. The independent service providers (ISPs) utilize the meter data to provide optional services offering the methods to save electricity usage through service modules. The energy suppliers are defined as commercial parties which generate or purchase electricity and sell it to customers. Grid operators typically use the information provided by the metering system, allocating electricity use.

The smart metering system features numerous ports through which information can pass to facilitate communication between system components and market parties. The NTA 8130 standard specifies four ports: P1, P2, P3, and P4 (as shown in Fig. 2).

Port P1 is the read-only port to connect metering installation with external devices. Port P2 is linked to other local metering apparatus. It can be used for wired or wireless connection with electricity/water/heating/hot water meters. Port P3 is used for two-way communication with distribution system operator (DSO) to send information such as metering values, status, power quality, and outage measurements. Communication between port P3 and DSO is achieved via long-term evolution, code-division multiple access, or packet radio service. Also, it uses the international standard IEC 62056 as the communication protocol [20]. Port P4 is a gateway for ISPs, energy suppliers, and grid operators to acquire measurements from port P3. It provides a web service to access the CAS, which allows an energy supplier or ISP to get customers' metering data, regardless of the responsible DSO.

Various kinds of services can be offered to customers by using smart metering system, such as generating remotely readable meter data, facilitating energy savings for consumers, and monitoring distribution networks. Also, a two-way communication network is essential for the functionalities of smart metering system. The exchange of messages, which contain the information about the status of meter installation and operating environments, is carried out between grid operators, energy suppliers, service providers, and consumers. For example, energy suppliers and grid operators can display current state information on the metering installation via port P3.

III. ANOMALY DETECTION USING BiLSTM AUTOENCODER

Power systems are being confronted with enormous information issues, and to deal with this huge volume of information, deep learning algorithms are the best arrangements [21]. In deep learning systems, different handling layers are considered and they may precisely extract the essential aspects of data with multiple levels of abstraction [22].

Recurrent neural network (RNN), long short-term memory (LSTM) network used for short-term prediction of power consumption [23] as an aid to replace multi-objective optimization of power consumption in [24], and gated recurrent unit (GRU) network applied for electricity trading [25] represent sorts of nonlinear predictor functions. Long-term dependencies are difficult for RNNs to learn, and the LSTM network is especially useful for time-series data. The LSTM network can accomplish better characterization of time-series data. The uni-directional LSTM approach produces certain errors in the results as they utilize one-way memory [26-28]. The BiLSTM approach is an extension of the traditional LSTM approach, which can improve the performance in the aspect of sequence classification by using the sequence information in both directions forward (past to future) and backward (future to past). The BiLSTM network has emerged as a better solution to deal with the interaction between contexts [29]. It was used to diagnose an asynchronous motor and found to be more efficient than the GRU network [30]. In issues where all timesteps of the information grouping are accessible, training the BiLSTM network represents training two uni-directional LSTM networks rather than one LSTM network on the input sequence. One LSTM network is trained by the input sequence and the other LSTM network is trained by a reversed copy of the input sequence. This can give the network more context and help it learn the problem faster and more thoroughly.

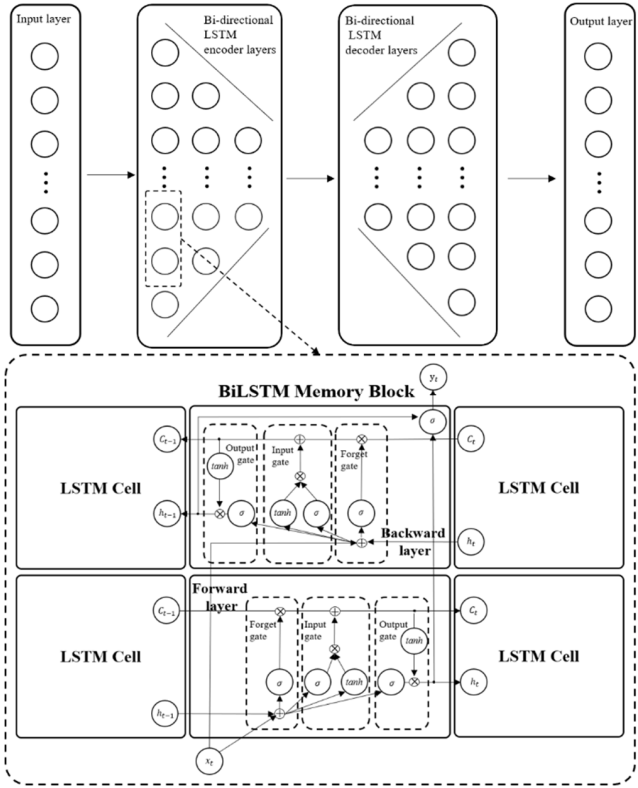

Fig. 3. Strucutre of BiLSTM autoencoder.

As shown in Fig. 3, an LSTM cell has forget, input, and output gates for effective short and long term cell memory. In particular, the training of BiLSTM memory block has two ways, one LSTM from past to future and the other one from future to past, possessing data from both the past and the future at each time. Each layer calculates the following function for each element in the input order. Variables involving with an LSTM cell are represented as following

$$i_t = \sigma(W_{ii}x_t + b_{ii} + W_{hi}h_{(t-1)} + b_{hi}) \quad (1)$$

$$f_t = \sigma(W_{if}x_t + b_{if} + W_{hf}h_{(t-1)} + b_{hf}) \quad (2)$$

$$g_t = \tanh(W_{ig}x_t + b_{ig} + W_{hg}h_{(t-1)} + b_{hg}) \quad (3)$$

$$o_t = \sigma(W_{io}x_t + b_{io} + W_{ho}h_{(t-1)} + b_{ho}) \quad (4)$$

$$c_t = f_t * c_{(t-1)} + i_t * g_t \quad (5)$$

$$h_t = o_t * \tanh(c_t) \quad (6)$$

where $(W_{ii}, W_{if}, W_{ig}, W_{io})$ are input weights, $(W_{hi}, W_{hf}, W_{hg}, W_{ho})$ are recurrent weights, $(b_{hi}, b_{hf}, b_{hg}, b_{ho})$ are recurrent biases, $h_t$ represents the hidden state at time $t$, $c_t$ depicts the cell state at time $t$, $x_t$ plays for the input at time $t$, $h_{(t-1)}$ is the hidden state of the layer at time $(t-1)$ or the initial hidden state at time 0, and $i_t$, $f_t$, $g_t$, $o_t$ represent the input, forget, cell, and output gate, respectively. $\sigma$ is the sigmoid function, and $'*'$ is the Hadamard product.

In most cases, the input data contains redundant or correlated characteristics, resulting in wasted processing time and model overfitting. An autoencoder comprises an encoder and a decoder. The encoder creates a compressed vector representation of the input data, which is then used by the decoder to create the target sequence. A pair of encoder and decoder is trained to reconstruct typical behavior, and then the reconstruction error is evaluated to detect anomalies. The input order is pierced successively in the LSTM autoencoder's structure, and after the last input sequence appears in the input-output sequence, the decoder restores the input sequence or outputs an estimated target sequence. In model learning, normal data is utilized instead of aberrant data, and the reconstruction error is measured by comparing the output with the input. The reconstruction error is obtained as follows

$$e_t = ||x_t - \hat{x}_t||^2$$
$$= \left|\left|x_t - \sigma(W_{io}x_t + b_{io} + W_{ho}h_{(t-1)} + b_{ho})\right|\right|^2 \quad (7)$$

$$O_{AD}(t) = \begin{cases} +1 \ (normal) & e_t \leq \theta \\ 0 \ (abnormal) & e_t > \theta \end{cases} \quad (8)$$

where $e_t$ represents the reconstruction error, $x_t$ and $\hat{x}_t$ are vectors of elements of 4 sorts of inputs and outputs of energy sources (electricity/water/heating/hot water), $O_{AD}(t)$ denotes anomaly detection parameter at $t$, $t$ is taken at every 15 minute interval, and $\theta$ is a threshold value for reconstruction error, and "abnormal" represents anomaly. When data is provided to the model, the $e_t$ is utilized to determine whether it is normal or abnormal by comparing $e_t$ with threshold $\theta$. If $e_t$ is less than the threshold, data is classified as normal, however, if $e_t$ is larger than threshold, data is classified as abnormal.

For the data preprocessing, data cleaning and examining, data denoising, and data normalization are employed for training the BiLSTM network. Classification and cleaning of 4 types of energy source data for each household, cleaning of tag values, and resampling at 15-minute intervals are carried out in data cleaning and sampling process. Data denoising with discrete wavelet transform (DWT) can also be utilized. The DWT method is used to effectively remove noise that is the cause of reduced performance of the trained BiLSTM network.

IV. SIMULATION RESULTS

In the smart metering system, data of electricity/water/heating/hot water are collected from 985 households. The training progress of the BiLSTM network is shown in Fig. 4 and its training parameters are as follows. The model is trained for 200 epochs with a specific batch size, using the Adam optimizer as a stochastic gradient descent. Also, the network is trained with a batch size of 128. The training loss is calculated by applying a mean squared error (MSE) as a loss function. To prevent inadequate learning such as local minimum and overfitting, dropout and gradient are applied.

Depending on the number of epochs, the decrease in both training loss and validation loss describes that BiLSTM autoencoder is well trained.

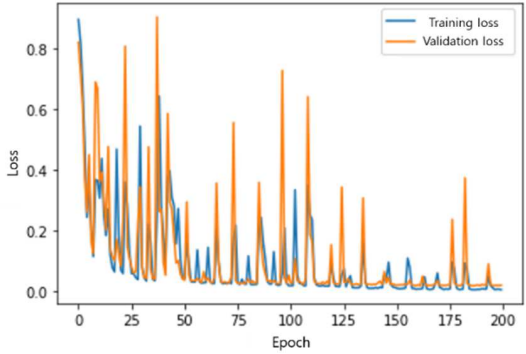

Fig. 4. Training progress of BiLSTM autoencoder.

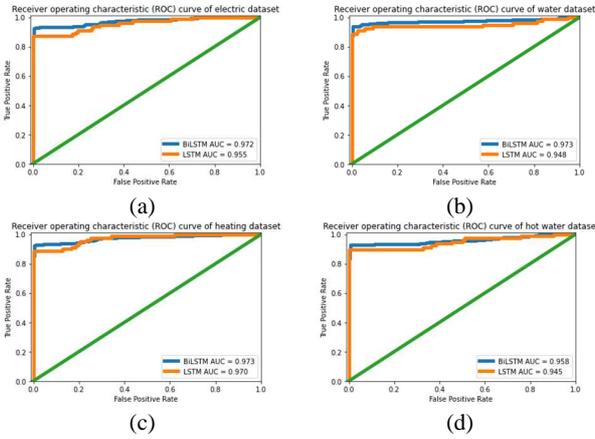

Fig. 5. ROC curve analysis in 4 types of energy sources: ROC curve of (a) electricity dataset; (b) water dataset; (c) heating dataset; (d) hot water dataset.

Table 1. Performance evaluation of uni-directional LSTM autoencoder and BiLSTM autoencoder.

| METHOD | ACCURACY | PRECISION | RECALL | $F_1$ | MSE |
|---|---|---|---|---|---|
| UNI-DIRECTIONAL LSTM AUTOENCODER | 0.99297 | 0.99313 | 0.99980 | 0.99646 | 0.34 |
| BI-DIRECTIONAL LSTM AUTOENCODER | 0.99575 | 0.99584 | 0.99988 | 0.99785 | 0.32 |

For the training of the BiLSTM autoencoder, 80% of data samples are used for training and the remaining 20% of data samples are used for validation. The receiver operating characteristic (ROC) curve is able to provide information on the overall performance of the classifier. As shown in Fig. 5, the BiLSTM autoencoder has larger area under curve (AUC) values than uni-directional LSTM autoencoder (represented by "LSTM" in the legend) with 4 types of energy sources electricity/water/heating/hot water. This demonstrates that the BiLSTM autoencoder achieves better performance in anomaly detection than the uni-directional LSTM autoencoder. Table 1 shows the accuracy, precision, recall, F1, and MSE defining the confusion matrix. Confusion matrix is composed of accuracy=(TP+TN)/(TP+FN+FP+TN), precision=TP/(TP+FP), recall=TP/(TP+FN), and $F_1$=2*precision*recall/(precision+recall). The TP, TN, FP, and FN denote true positive, true negative, false positive and false negative in classification results.

As shown in Fig. 6, anomalies are detected in 4 types of energy sources with the BiLSTM autoencoder by calculating the reconstruction error between input and reconstructed input. The level of outliers for each data point is described by the anomaly score. A domain-specific threshold is chosen by a subject matter expert to identify the anomalies, and the data instances are sorted according to anomalous score. Dark blue dot, orange dot, and thin red line represent normal data, outlier, and threshold. When data points have higher reconstruction error than the threshold, they are classified as outlier (anomaly). On the contrary, when data points have less reconstruction error than the threshold, they are classified as normal data. There exist other outliers which are classified as normal data because of small reconstruction error similarly observed with normal data and some normal data classified as outliers due to the reconstruction error above the threshold.

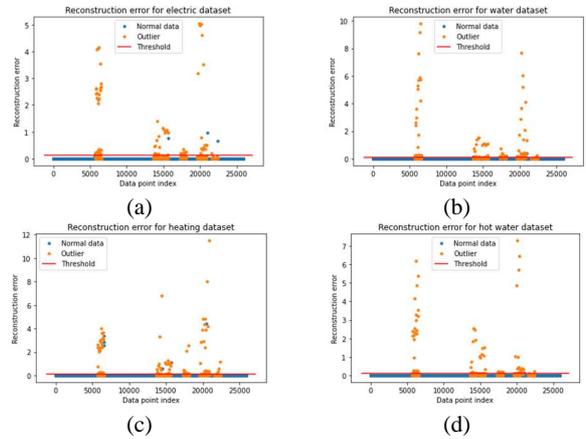

Fig. 6. Anomaly detection in 4 types of energy sources: anomaly detection of (a) electricity dataset; (b) water dataset; (c) heating dataset; (d) hot water dataset.

## V. CONCLUSION

In this paper, a novel method of anomaly detection for smart metering system is proposed. The dataset, which is composed of electricity, water, heating, and hot water and collected from 985 households, is used for simulations. To find anomalies among datasets measured by smart metering system, the BiLSTM autoencoder is utilized. To prove the comparative performance of the BiLSTM autoencoder in anomaly detection, ROC curve and AUC are calculated. As presented in Fig. 5, the BiLSTM autoencoder achieves better performance, as compared to the uni-directional LSTM autoencoder, for 4 types of energy sources due to the structure of the BiLSTM network making the input flow in forward and backward directions to preserve the past and the future information.

ACKNOWLEDGMENTS

This work was supported by the Korea Institute of Energy Technology Evaluation and Planning (KETEP) and the Ministry of Trade, Industry & Energy (MOTIE) of the Republic of Korea (No. 20192010107290).

REFERENCES

[1] A. Goulding, "A New Blueprint for Ontario's Electricity Market," *Commentary-CD Howe Institute,* no. 389, p. 0_1, 2013.
[2] J. Zheng, D. W. Gao, and L. Lin, "Smart meters in smart grid: An overview," in *2013 IEEE Green Technologies Conference (GreenTech)*, 2013: IEEE, pp. 57-64.
[3] G. R. Barai, S. Krishnan, and B. Venkatesh, "Smart metering and functionalities of smart meters in smart grid-a review," in *2015 IEEE Electrical Power and Energy Conference (EPEC)*, 2015: IEEE, pp. 138-145.
[4] N. M. G. Strategy, "Advanced metering infrastructure," *US Department of Energy Office of Electricity and Energy Reliability,* 2008.
[5] A. Snyder and M. G. Stuber, "The ANSI C12 protocol suite-updated and now with network capabilities," in *2007 Power Systems Conference: Advanced Metering, Protection, Control, Communication, and Distributed Resources*, 2007: IEEE, pp. 117-122.
[6] M. Öhrström, "Fast fault detection for power distribution systems," Elektrotekniska system, 2003.
[7] H. Tram, "Technical and operation considerations in using smart metering for outage management," in *2008 IEEE/PES Transmission and Distribution Conference and Exposition*, 2008: IEEE, pp. 1-3.
[8] E. Hong, K. Kim, and D. Har, "Spectrum sensing by parallel pairs of cross-correlators and comb filters for OFDM systems with pilot tones," *IEEE Sensors Journal,* vol. 12, no. 7, pp. 2380-2383, 2012.
[9] H. Kim, E. Hong, C. Ahn, and D. Har, "A pilot symbol pattern enabling data recovery without side information in PTS-based OFDM systems," *IEEE Transactions on Broadcasting,* vol. 57, no. 2, pp. 307-312, 2011.
[10] C. Moraes and D. Har, "Charging distributed sensor nodes exploiting clustering and energy trading," *IEEE Sensors Journal,* vol. 17, no. 2, pp. 546-555, 2016.
[11] C. Moraes, S. Myung, S. Lee, and D. Har, "Distributed sensor nodes charged by mobile charger with directional antenna and by energy trading for balancing," *Sensors,* vol. 17, no. 1, p. 122, 2017.
[12] D. M. Hawkins, *Identification of outliers*. Springer, 1980.
[13] Y. Heo, R. Choudhary, and G. Augenbroe, "Calibration of building energy models for retrofit analysis under uncertainty," *Energy and Buildings,* vol. 47, pp. 550-560, 2012.
[14] P. Malhotra, A. Ramakrishnan, G. Anand, L. Vig, P. Agarwal, and G. Shroff, "LSTM-based encoder-decoder for multi-sensor anomaly detection," *arXiv preprint arXiv:1607.00148,* 2016.
[15] C. Fan, F. Xiao, Y. Zhao, and J. Wang, "Analytical investigation of autoencoder-based methods for unsupervised anomaly detection in building energy data," *Applied energy,* vol. 211, pp. 1123-1135, 2018.
[16] S. Lv, J. Wang, Y. Yang, and J. Liu, "Intrusion prediction with system-call sequence-to-sequence model," *IEEE Access,* vol. 6, pp. 71413-71421, 2018.
[17] S.-V. Oprea, A. Bâra, F. C. Puican, and I. C. Radu, "Anomaly Detection with Machine Learning Algorithms and Big Data in Electricity Consumption," *Sustainability,* vol. 13, no. 19, p. 10963, 2021.
[18] M. Ismail, M. F. Shaaban, M. Naidu, and E. Serpedin, "Deep learning detection of electricity theft cyber-attacks in renewable distributed generation," *IEEE Transactions on Smart Grid,* vol. 11, no. 4, pp. 3428-3437, 2020.
[19] "Sustainable Buildings - Smart Meters, 2008, http://w1.siemens.com/innovation/en/publikationen/publications-pof/pof-fall-2008/gebaeude.htm."
[20] I. E. Commission, "Electricity metering data exchange—the DLMS/COSEM suite—part 5-3: DLMS/COSEM application layer," IEC 62056-5-3: 2016, ed, 2016.
[21] N. Rusk, "Deep learning," *Nature Methods,* vol. 13, no. 1, pp. 35-35, 2016.
[22] T. Kim, L. F. Vecchietti, K. Choi, S. Lee, and D. Har, "Machine Learning for Advanced Wireless Sensor Networks: A Review," *IEEE Sensors Journal,* 2020.
[23] S. Lee, H. Jin, L. F. Vecchietti, J. Hong, and D. Har, "Short-term predictive power management of PV-powered nanogrids," *IEEE Access,* vol. 8, pp. 147839-147857, 2020.
[24] S. Lee *et al.*, "Optimal power management for nanogrids based on technical information of electric appliances," *Energy and Buildings,* vol. 191, pp. 174-186, 2019.
[25] S. Lee *et al.*, "Cooperative decentralized peer‐to‐peer electricity trading of nanogrid clusters based on predictions of load demand and PV power generation using a gated recurrent unit model," *IET Renewable Power Generation,* 2021.
[26] L. Wen, K. Zhou, S. Yang, and X. Lu, "Optimal load dispatch of community microgrid with deep learning based solar power and load forecasting," *Energy,* vol. 171, pp. 1053-1065, 2019.
[27] S. Wen, Y. Wang, Y. Tang, Y. Xu, P. Li, and T. Zhao, "Real-time identification of power fluctuations based on lstm recurrent neural network: A case study on singapore power system," *IEEE Transactions on Industrial Informatics,* vol. 15, no. 9, pp. 5266-5275, 2019.
[28] Y. Zhou, F.-J. Chang, L.-C. Chang, I.-F. Kao, and Y.-S. Wang, "Explore a deep learning multi-output neural network for regional multi-step-ahead air quality forecasts," *Journal of cleaner production,* vol. 209, pp. 134-145, 2019.
[29] M. Schuster and K. K. Paliwal, "Bidirectional recurrent neural networks," *IEEE transactions on Signal Processing,* vol. 45, no. 11, pp. 2673-2681, 1997.
[30] N. Enshaei and F. Naderkhani, "Application of deep learning for fault diagnostic in induction machine's bearings," in *2019 IEEE International Conference on Prognostics and Health Management (ICPHM)*, 2019: IEEE, pp. 1-7.